\documentclass[sigconf,nonacm]{acmart}

\renewcommand\footnotetextcopyrightpermission[1]{} 

\AtBeginDocument{%
  }

\usepackage{amsmath, amssymb}
\usepackage{amsthm}
\usepackage{graphicx}
\usepackage{algorithm}
\usepackage{algpseudocode}
\usepackage{hyperref}

\usepackage{enumitem}
\usepackage{algorithm, algpseudocode} 
\usepackage{tikz} 
\usepackage{hyperref} 

\newcommand{\dd}{\mathrm{d}}

\newcommand{\E}{\mathbb{E}}

\newcommand{\g}{\mathcal{G}}

\newtheorem{theorem}{Theorem}

\newtheorem{remark}{Remark}

\algblock{Input}{EndInput}
\algnotext{EndInput}
\algblock{Output}{EndOutput}
\algnotext{EndOutput}

\usetikzlibrary{fit,positioning}

\newcommand{\bm}[1]{\boldsymbol{#1}}

\begin{document}


\title{Continuous-Time Reinforcement Learning for Asset–Liability Management}



\author{Yilie Huang}
\email{yh2971@columbia.edu}
\orcid{0000-0002-2922-9904}
\affiliation{%
  \institution{Department of Industrial Engineering and Operations Research, Columbia University}
  \city{New York}
  \state{New York}
  \country{USA}
}

\renewcommand{\shortauthors}{}

\begin{abstract}
   This paper proposes a novel approach for Asset–Liability Management (ALM) by employing continuous-time Reinforcement Learning (RL) with a linear-quadratic (LQ) formulation that incorporates both interim and terminal objectives. We develop a model-free, policy gradient-based soft actor-critic algorithm tailored to ALM for dynamically synchronizing assets and liabilities. To ensure an effective balance between exploration and exploitation with minimal tuning, we introduce adaptive exploration for the actor and scheduled exploration for the critic. Our empirical study evaluates this approach against two enhanced traditional financial strategies, a model-based continuous-time RL method, and three state-of-the-art RL algorithms. Evaluated across 200 randomized market scenarios, our method achieves higher average rewards than all alternative strategies, with rapid initial gains and sustained superior performance. The outperformance stems not from complex neural networks or improved parameter estimation, but from directly learning the optimal ALM strategy without learning the environment.
\end{abstract}

\keywords{Continuous-time Reinforcement Learning, Asset-Liability Management, Model-Free, Actor-Critic}

\titlenote{Published in the \textit{Proceedings of the 6th ACM International Conference on AI in Finance} (ICAIF 2025), pp.\ 360--368. \url{https://doi.org/10.1145/3768292.3770337}.}
\maketitle

\section{Introduction}
\label{sec:intro}

Asset-Liability Management (ALM) \citep{van1993strategies} is a critical component of financial strategy, involving the careful coordination of assets and liabilities to ensure the financial health of institutions. It plays a crucial role for banks, insurance companies, and pension funds, where the alignment of assets against liabilities significantly affects financial stability and regulatory compliance. 

Traditionally, ALM has utilized a range of methods to effectively synchronize assets and liabilities: static approaches like cash flow matching \citep{wise1984matching} ensure liabilities are met with corresponding asset inflows, and passive value-driven strategies such as key rate duration matching \citep{ho1992key} mitigate interest rate fluctuations by aligning the durations of assets and liabilities. On the other hand, more dynamic techniques, such as contingent immunization \citep{leibowitz1982contingent,leibowitz1983contingent} and Constant Proportion Portfolio Insurance (CPPI) \citep{hakanoglu1989constant,black1992theory}, actively adjust asset allocation to maintain a target surplus or minimize deviation from a predefined target. Maintaining a target surplus is essential for balancing solvency and efficient capital use, helping to mitigate both the risk of insolvency from insufficient surplus and the inefficiency of excess capital. However, these traditional methods often assume a stable environment with complete information, which limits their adaptability in fast-changing market conditions.

Reinforcement learning (RL) offers notable advantages in ALM by dynamically adjusting policies based on real-time feedback, making it a powerful tool for decision-making in dynamic and uncertain environments. Despite its potential, most RL methods have been developed around discrete-time Markov decision processes (MDPs) with discrete state and action spaces. While effective in many domains, this discrete-time framework faces inherent limitations when applied to systems that naturally evolve in continuous time, such as financial markets. Bridging this gap requires discretizing continuous-time problems into discrete-time models, enabling the use of standard RL algorithms. However, this discretization introduces critical challenges. Selecting an appropriate time step size is particularly challenging: large time steps can oversimplify the problem, reducing resolution and yielding suboptimal policies, while small time steps, despite their precision, increase the computational burden and may result in instability \citep{munos2006policy,tallec2019making,park2021time}. Additionally, discretization often struggles to accurately capture complex, fine-grained dynamics in continuous environments, especially in high-frequency decision-making contexts. This mismatch between discrete-time models and continuous-time dynamics limits the effectiveness of traditional RL methods in such settings, underscoring the need for continuous-time RL frameworks that can better handle the inherent complexities of financial markets.

Recent advancements in continuous-time RL have marked a pivotal shift from earlier isolated studies \citep{baird1994reinforcement,doya2000reinforcement,vamvoudakis2010online} to a more cohesive and systematic framework. The introduction of an entropy-regularized control approach by \cite{wang2020reinforcement} laid a rigorous mathematical foundation for this field. Building on this, subsequent research \citep{jia2021policy,jia2021policypg} developed comprehensive methods for policy evaluation and improvement. A key aspect of this research is its model-free, data-driven approach, which focuses on directly learning optimal control policies without requiring explicit model estimation. This progress has not only solidified the theoretical underpinnings of continuous-time RL but also inspired diverse extensions and practical applications in domains requiring robust real-time decision-making under uncertainty \citep{huang2022achieving,tang2024regret,huang2024mean,wei2023continuous}.

While recent developments in continuous-time RL have made significant theoretical progress, much of the existing work remains primarily focused on analytical results rather than empirical validation. To date, these methods have not been applied to ALM, a setting that naturally fits the continuous-time framework due to its dynamic and stochastic nature. Furthermore, empirical comparisons between continuous-time RL and traditional discrete-time RL approaches in financial applications remain largely unexplored, leaving open the question of their relative practical effectiveness.

Building upon these theoretical advancements, our work mainly contributes in the following ways:
\begin{itemize}
    \item A linear-quadratic (LQ) formulation of the ALM problem is introduced, incorporating both interim and terminal objectives, and addressing the time-inconsistency and terminal constraint issues in the traditional mean-variance (MV) approach.
    \item We design a model-free, policy gradient-based (soft) actor--critic algorithm tailored for the ALM problem within a continuous-time RL framework. To the best of our knowledge, this is the first work that applies continuous-time RL to ALM.
    \item We introduce an adaptive exploration mechanism for the actor and a scheduled exploration strategy for the critic, enabling a robust exploration--exploitation trade-off throughout learning. 
    \item Empirical results demonstrate that our algorithm achieves superior performance over traditional ALM strategies, model-based continuous-time RL, and advanced RL baselines.
\end{itemize}

The remainder of the paper is organized as follows. Section \ref{sec:ALM} formulates the ALM problem and discusses preliminary results essential for subsequent developments. Section \ref{sec:algorithm} outlines our RL algorithm's development and principles. Section \ref{sec:theorem} proves the convergence of the proposed algorithm. Section \ref{sec:numerical_experiments} evaluates our algorithm against six ALM strategies and presents experimental results. Finally, Section \ref{sec:conclusions} concludes.

\section{Description of the Asset-Liability Management Problem}
\label{sec:ALM}

\subsection{Classical Stochastic LQ Framework for ALM}
\label{sec:stochastic_control_framework}

One widely used approach to address the ALM problem is through stochastic control, particularly using the MV formulation, as demonstrated in several studies \citep{shen2020mean,hu2022non,pan2018optimal}. In this framework, the state variable is typically defined as the surplus, representing the difference between assets and liabilities. The primary objective is to minimize the expected squared deviation of the terminal surplus from a predefined target surplus, subject to a terminal constraint.

While the MV formulation provides a well-structured framework for ALM, it also presents several key limitations. First, MV-based formulations suffer from an inherent time-inconsistency issue, meaning that strategies deemed optimal at the outset may become suboptimal as time progresses \citep{zhou2000continuous}. Second, addressing the terminal constraint in MV formulations often requires additional techniques, such as introducing Lagrange multipliers, which can complicate the solution. Finally, by focusing exclusively on minimizing the terminal surplus deviation, the MV approach overlooks the importance of managing the surplus throughout the entire horizon, which is crucial for ensuring financial stability and meeting ongoing obligations.

To address these limitations, we formulate the ALM problem as a specific stochastic LQ control problem, which is structurally similar to the MV approach but introduces key differences. The first distinction lies in the state representation: instead of directly modeling the surplus, we define the state variable \(x(t) \in \mathbb{R}\) as the surplus deviation, representing the difference between the surplus and the target surplus:
\[
x(t) = \text{Assets}(t) - \text{Liabilities}(t) - \text{Target Surplus},
\]
where a positive \( x(t) \) indicates a surplus above the target, implying inefficient capital use, while a negative \( x(t) \) represents a shortfall, increasing the risk of insolvency.

By modeling the deviation directly, this formulation simplifies the ALM problem and provides a clearer objective aligned with maintaining a stable surplus relative to a predefined target. It also allows for more precise tracking of the financial position by explicitly focusing on deviations rather than the raw surplus deviation. 

The control variable \( u(t) \in \mathbb{R}\) represents strategic financial decisions, such as asset reallocation, funding adjustments, and liability management, aimed at minimizing deviations from the target surplus over time.

The dynamics of surplus deviation \( x \) under the influence of financial control \( u \) follow the stochastic differential equation (SDE):  
\begin{equation}
\label{eq_classical_dynamics}
\mathrm{d}x^u(t) = (A x^u(t) + B u(t)) \mathrm{d}t + (C x^u(t) + D u(t)) \mathrm{d}W(t),
\end{equation}
where \(x^u(0) = x_0\) is the initial surplus deviation and \(W(t)\) is standard Brownian motion, representing the stochastic nature of financial markets. The model parameters are interpreted as follows:

\begin{itemize}
    \item \(A\): Represents the internal drift, modeling the natural tendency of the surplus deviation to increase or decrease over time without intervention.
    \item \(B\): Captures the direct impact of the control \( u \) on the surplus deviation. A larger \( B \) means control actions have a stronger influence.
    \item \(C\): Scales the impact of the current surplus deviation on the volatility of the system. Higher values of \(C\) amplify how fluctuations in the surplus deviation contribute to uncertainty in the dynamics.
    \item \(D\): Describes how control actions affect the variability of the surplus deviation. Higher values of \(D\) imply that control actions have a stronger impact on the uncertainty in the system.
\end{itemize}

Our objective is to manage \( u \) strategically to minimize deviations from the target surplus, penalizing both positive and negative deviations. Positive deviations imply surplus accumulation beyond the target, which could result in inefficient capital use or reduced financial efficiency. Negative deviations, on the other hand, indicate a shortfall relative to the target, reducing the safety buffer against financial uncertainties. We aim to optimize the expected value of the quadratic objective functional that incorporates both interim and terminal deviations:
\begin{equation}
\label{eq_classical_lq}
    \max_{u} \mathbb{E} \left[ \int_0^T -\frac{1}{2} Q x^u(t)^2 \mathrm{d}t - \frac{1}{2} H x^u(T)^2 \right],
\end{equation}
where \(Q \geq 0\) and \(H \geq 0\) are coefficients that penalize deviations from the target surplus over the time horizon \([0, T]\) and at the terminal time \(T\), respectively. 

Notably, unlike conventional formulations, we do not explicitly penalize the control \( u \) in the objective function. Instead, \( u \) is implicitly constrained through the dynamics given in \eqref{eq_classical_dynamics}. This type of formulation has led to an active research area known as ``indefinite stochastic Linear Quadratic control'' \citep{chen1998stochastic,rami2000linear}.

Provided the model parameters \(A\), \(B\), \(C\), \(D\), \(Q\), and \(H\) are known, the established stochastic control theory can solve this optimization problem \citep{YZbook}, producing an optimal value function and control policy:
\begin{equation}
\label{eq_classical_value_function}
\begin{aligned}
    V^{CL}(t,x) &= -\frac{1}{2} \left[ \frac{Q}{\Lambda} + (H-\frac{Q}{\Lambda}) e^{\Lambda (t-T)} \right] x^2, \\
    u^{CL}(t,x) &= -\frac{B+CD}{D^2}x,
\end{aligned}
\end{equation}
where \(\Lambda = \frac{1}{D^2} (B^2 + 2BCD - 2AD^2)\).

\subsection{Continuous-Time RL Framework for ALM}

However, the complete and precise knowledge of parameters such as \(A\), \(B\), \(C\), and \(D\) in real-life ALM scenarios is often impractical, necessitating the use of RL to manage uncertainties. RL addresses these challenges by maintaining a balance between exploration and exploitation, adapting dynamically to the unknown parameters of the environment \citep{sutton2018reinforcement}. This is achieved through randomized control processes, where controls \(u\) are derived from a distribution \(\pi=\{\pi(\cdot,t) \in \mathcal{P}(\mathbb{R}) : 0 \leq t \leq T\}\), representing all probability density functions over \(\mathbb{R}\). To encourage exploration, an entropy term is integrated into the objective function, promoting stochastic policies. This approach is conceptually related to soft-max approximations and Boltzmann exploration strategies \citep{haarnoja2018soft,ziebart2008maximum}.

By \cite{wang2020reinforcement}, under entropy-regularized RL for continuous-time controlled diffusion processes, the dynamics of the ALM problem under stochastic policy \(\pi\) are given by:
\begin{equation}
\mathrm{d}x^\pi(t) = \widetilde{b}(x^\pi(t),\pi(\cdot,t)) \mathrm{d}t + \widetilde{\sigma}(x^\pi(t),\pi(\cdot,t)) \mathrm{d}W(t),
\label{rl_dynamics}
\end{equation}
where the drift \(\widetilde{b}\) and diffusion \(\widetilde{\sigma}\) components are defined as:
\begin{equation}
\begin{aligned}
\widetilde{b}(x,\pi) &= A x + B \int_{\mathbb{R}} u \pi(u) \mathrm{d}u, \\
\widetilde{\sigma}(x,\pi) &= \sqrt{\int_{\mathbb{R}} (C x + D u)^2 \pi(u) \mathrm{d}u}, \quad(x,\pi)\in \mathbb{R}\times \mathcal{P}(\mathbb{R}).
\end{aligned}
\label{b_sigma_tilde}
\end{equation}

The entropy-regularized value function for stochastic policy \(\pi\) is expressed as:
\begin{equation}
\label{eq_value_function_pi}
\begin{aligned}
    J(t, x; \pi) = \mathbb{E} \biggl[ \int_t^T \biggl(&-\frac{1}{2}Qx^\pi(s)^2 + \gamma p(s) \biggl) \mathrm{d}s \\
    &- \frac{1}{2}Hx^\pi(T)^2 \Big| x^\pi(t)=x \biggl],
\end{aligned}
\end{equation}
where \(p(t) = -\int_{\mathbb{R}} \pi(t,u) \log \pi(t,u) \mathrm{d}u\) represents the entropy term, and \(\gamma\), known as the temperature parameter, is the weight on exploration.

The optimal value function and optimal randomized/stochastic (feedback) policy are solved as follows:
\begin{equation}
\label{eq_rl_value_function}
\begin{aligned}
    V(t, x) &= -\frac{1}{2}k_1(t)x^2 + k_3(t), \\
   \pi(u \mid t, x) &= \mathcal{N}\left(u \Big | -\frac{(B+CD)}{D^2}x, \frac{\gamma}{D^2 k_1(t)}\right),
\end{aligned}
\end{equation}
where $k_1>0$ and $k_3$ are certain functions of $t$ that can be determined completely by the model parameters. 

It should be noted that the values of all model parameters are unknown to the agent, meaning that the optimal solutions in \eqref{eq_rl_value_function} cannot be directly applied. Moreover, we make no attempt to estimate these parameters, as is typically done in model-based approaches. Instead, we adopt a model-free approach that entirely avoids model estimation. Despite the unknown parameters, this model provides critical insights into the structural properties of the optimal solutions, thereby reducing the complexity of function parameterization and approximation in the learning process. This advantage will be demonstrated in the next section.

\section{A Continuous-Time RL Algorithm}
\label{sec:algorithm}
This section presents a continuous-time RL algorithm specifically designed for ALM. It covers critical aspects including function parameterization, policy evaluation and improvement methods, adaptive actor exploration, and scheduled critic exploration. Finally, we provide discretized updating rules and pseudocode for our ALM-RL algorithm.

\subsection{Function Parameterization}
\label{sec:parameterization}

While the direct application of the optimal solutions \eqref{eq_rl_value_function} from the continuous-time RL framework is impractical due to unknown model parameters, the structural insights guide our parameterization. Specifically, the optimal value function is quadratic in the surplus deviation \(x\), and the mean of the optimal stochastic Gaussian policy is linearly dependent on \(x\). Thus, we parameterize the value function with parameters \(\bm{\theta} \in \mathbb{R}^d\):

\begin{equation}
\label{value_parametrization}
    J(t, x; \bm{\theta}) = -\frac{1}{2}k_1(t; \bm{\theta})x^2 + k_3(t; \bm{\theta}),
\end{equation}
where both functions \(k_1\) and \(k_3\) are continuous in \(t\) and \(\bm\theta\). And the policy with \(\bm{\phi} = (\phi_1, \phi_2 > 0)^\top\), yielding a Gaussian distribution:
\begin{equation}
\label{policy_parametrization}
\pi(u \mid x; \bm{\phi}) = \mathcal{N}(u \mid \phi_1 x, \phi_2).
\end{equation}

\subsection{Policy Evaluation}
\label{sec:policy_evaluation}
Policy evaluation (PE) is a critical component in RL, focusing on learning the value function associated with a given control policy. Following the parameterization strategies outlined for the value function and policy in \eqref{value_parametrization} and \eqref{policy_parametrization}, PE involves updating the parameters \(\bm{\theta}\) to refine the function approximations of \(k_1(t; \bm{\theta})\) and \(k_3(t; \bm{\theta})\). The Temporal Difference (TD) method proposed in \cite{jia2021policy} suggests an offline learning setting for updating \(\bm{\theta}\) as follows:

\begin{equation}
\begin{aligned}
\label{eq_theta_update_without_projection}
\bm\theta_{n+1} \leftarrow  \bm\theta_n + a_n\int_0^T &\frac{\partial J}{\partial \bm\theta}(t,x_n(t);\bm\theta_n)  \biggl[\mathrm{d} J\left(t, x_n(t); \bm\theta_n \right) \\
&- \frac{1}{2}Qx_n(t)^2 \dd t + \gamma {p}\left(t, \bm\phi_n\right) \mathrm{d} t\biggl],
\end{aligned}
\end{equation}
where \(a_n\) denotes the learning rate, and the subscript \(n\) indicates the \(n\)-th episode throughout.

Furthermore, \cite{huang2024sublinear} theoretically proves that the convergence rate of policy parameters \(\bm{\phi}\) is robust to the forms of \(k_1(t; \bm{\theta})\) and \(k_3(t; \bm{\theta})\), enabling flexible adaptations across different complexities in ALM modeling.

\subsection{Policy Improvement}
\label{sec:policy_iteration}
Policy improvement enhances the policy by iteratively updating the policy parameters based on feedback from the environment, aiming to increase expected performance. We adopt the continuous-time policy gradient (PG) method from \cite{jia2021policypg} for \(\phi_1\). Moreover, to ensure stability when updating \(\phi_{1,n}\), we need to consider the effect of diminishing exploration controlled by \(\phi_{2,n}\). As \(\phi_{2,n}\) becomes small, the term \(\phi_{2,n}^{-1}\) appearing in \(\frac{\partial \log \pi}{\partial \phi_1}\) can lead to numerical instability. To address this, we multiply by \(\phi_{2,n}\) during the update, effectively neutralizing the inverse and stabilizing the learning process.

\begin{equation}
\label{eq_phi_updates_without_projections}
\phi_{1,n+1} \leftarrow \phi_{1,n} + a_n Z_{1,n}(T),
\end{equation}
where \(a_n\) is the learning rate and the term \(Z_{1,n}(s)\) is defined as:
\begin{equation}
\label{eq_z1n_def}
\begin{aligned}
Z_{1,n}(s) =& \int_{0}^{s}  \frac{\partial \log \pi}{\partial \phi_1}\left(u_n(t) \mid x_n(t); \bm\phi_n\right) \biggl[\mathrm{d} J(t, x_n(t); \bm\theta_n) \\
&- \frac{1}{2}Q x_n(t)^2 \dd t + \gamma {p}(t, \bm\phi_n) \mathrm{d} t\biggl] \phi_{2,n}.
\end{aligned}
\end{equation}

\subsection{Adaptive Actor Exploration}
The actor’s exploration level is governed by the variance of the stochastic policy, represented by \(\phi_2\). In the approach proposed by \cite{huang2024sublinear}, \(\phi_{2,n}\) follows a predetermined diminishing sequence, limiting the adaptability of exploration to evolving data.

To improve the adaptability of actor exploration, we employ the policy-gradient updating method from \cite{huang2025data}, which enables \(\phi_2\) to be updated dynamically in response to observed data. For computational efficiency in the stochastic approximation algorithm, we reparametrize \(\phi_2\) as \(\phi_2^{-1}\). By applying the chain rule, the derivative of \(\phi_2^{-1}\) with respect to \(\phi_2\) simplifies to a time-invariant factor, which can be ignored in the gradient update, streamlining the computation. Consequently, we have

\begin{equation}
\label{lq2_eq_phi2_updates_without_projections}
\phi_{2, n+1} \leftarrow \phi_{2,n} - a_n Z_{2,n}(T),
\end{equation}
where $a_n$ is the learning rate, and \(Z_{2,n}(s)\) is defined as follows:
\begin{equation}
\label{lq2_eq_z2_def1}
\begin{aligned}
Z_{2,n}(s) =& \int_{0}^{s}\biggl\{\frac{\partial \log \pi}{\partial \phi_{2,n}^{-1}}\left(u_n(t) \mid t, x_n(t); \bm\phi_n\right)\\
&\biggl[\mathrm{d} J\left(t, x_n(t); \bm\theta_n \right) - \frac{1}{2}Qx_n(t)^2 \mathrm{d} t + \gamma p\left(t, \bm\phi_n\right) \mathrm{d} t\biggl] \\
&+ \gamma \frac{\partial p}{\partial \phi_{2,n}^{-1}}\left(t, \bm\phi_n \right) \mathrm{d} t \biggl\}.
\end{aligned}
\end{equation}

\subsection{Scheduled Critic Exploration}

The temperature parameter \(\gamma\) plays a crucial role in RL by controlling the weight of the entropy-regularized term in the objective function, as outlined in \eqref{eq_value_function_pi}. This parameter governs the level of exploration by the critic, influencing how much variability is incorporated into policy evaluations. A high \(\gamma\) value promotes exploration by emphasizing entropy, while a low \(\gamma\) value focuses on exploitation, which can lead to faster convergence but risks premature policy stagnation.

Maintaining a balance between exploration and exploitation is essential to ensure that the algorithm explores sufficiently during the early stages while converging effectively in later stages. To achieve this, \(\gamma\) needs to diminish over time, allowing the critic to gradually shift its focus from exploration to exploitation. Instead of using a fixed hyperparameter \(\gamma\) that requires extensive fine-tuning, as commonly done in continuous-time RL \citep{jia2021policy,jia2021policypg}, we propose a scheduled approach, where \(\gamma\) is defined as:

\begin{equation}
\label{lq2_eq_gamma_definition}
\gamma_n = \frac{c_\gamma}{b_n}, \qquad \text{for }n=0,1,\cdots
\end{equation}
where \(c_\gamma\) is a constant that determines the exploration level, while \(b_n > 1\) represents a monotone increasing sequence to infinity that governs the exploration scheduling. This formulation ensures that \(\gamma\) decreases systematically over time, providing a natural mechanism to balance exploration and exploitation without manual tuning.

\subsection{Discretization and Projections}
In our continuous-time RL framework, both the theoretical development and analysis are carried out entirely in continuous time. Discretization is introduced only at the final stage, solely for numerical implementation—specifically for approximating integrals and computing the \( \dd J \) term. To this end, the time interval \([0, T]\) is divided into uniform steps of length \(\Delta t\). This final-stage discretization avoids the drawbacks of discretizing the problem at the outset (i.e., converting the continuous-time problem into a Markov Decision Process), which is known to cause performance instability when the timestep is small \citep{munos2006policy,tallec2019making,park2021time}.

To ensure numerical stability during learning, we project the parameters onto convex sets:
\[
\begin{aligned}
K_{\bm\theta} &= \left\{ \bm\theta \in \mathbb{R}^d : |\bm\theta| \leq U_{\bm\theta} \right\}, \qquad K_{1} = \left\{ \phi_1 \in \mathbb{R} : |\phi_1| \leq U_{1} \right\}, \\
K_{2} &= \left\{ \phi_2 \in \mathbb{R} : \epsilon \leq |\phi_2| \leq U_{2} \right\},
\end{aligned}
\]
where \(U_{\bm\theta}\), \(U_{1}\), and \(U_{2}\) are fixed, sufficiently large positive constants that bound the parameter magnitudes. The constant \(\epsilon > 0\) represents the minimum exploration level to enforce non-degenerate stochastic policies and can be chosen arbitrarily close to zero. In practice, these bounds can be tuned to improve empirical performance while preserving theoretical stability.

Finally, for convex set $K$, we define $\Pi_{K}(x):=\arg \min_{y\in K} |y-x|^2$. By employing a scheduled temperature \(\gamma_n\) and applying Euler discretization, the update rules for the parameters \(\bm\theta_n\), \(\phi_{1,n}\), and \(\phi_{2,n}\) in \eqref{eq_theta_update_without_projection}, \eqref{eq_phi_updates_without_projections}, and \eqref{lq2_eq_phi2_updates_without_projections} are derived as follows:

\begin{equation}
\label{lq2_eq_theta_update_discrete}
\begin{aligned}
\bm\theta_{n+1}& \leftarrow \Pi_{K_{\bm\theta}}\biggl( \bm\theta_n+a_n \sum_{k=0}^{\left\lfloor \frac{T}{\Delta t} -1 \right\rfloor} 
\frac{\partial J}{\partial \bm\theta}(t_k,x_n(t_k);\bm\theta_n) \biggl[ - \frac{1}{2}Qx_n(t_k)^2 \Delta t \\
& + \gamma_n {p}\left(t_k, \bm\phi_n\right) \Delta t  + J\left(t_{k+1}, x_n(t_{k+1}); \bm\theta_n \right) - J\left(t_k, x_n(t_k); \bm\theta_n \right) \biggl] \biggr),
\end{aligned}
\end{equation}

\begin{equation}
\label{lq2_eq_phi1_update_discrete}
\begin{aligned}
\phi_{1, n+1}& \leftarrow \Pi_{K_{1}}\biggl( \phi_{1,n} + a_n \phi_{2,n} \sum_{k=0}^{\left\lfloor \frac{T}{\Delta t} -1 \right\rfloor} 
\biggl\{ \frac{\partial \log \pi}{\partial \phi_1}\left(u_n(t_k) \mid t_k, x_n(t_k); \bm\phi_n\right) \\
&\biggl[ J\left(t_{k+1}, x_n(t_{k+1}); \bm\theta_n \right) 
- J\left(t_k, x_n(t_k); \bm\theta_n \right) - \frac{1}{2}Qx_n(t_k)^2 \Delta t \\
&+ \gamma_n {p}\left(t_k, \bm\phi_n\right) \Delta t \biggl] + \gamma_n \frac{\partial {p}}{\partial \phi_1}\left(t_k, \bm\phi_n \right) \Delta t \biggl\} \biggr),
\end{aligned}
\end{equation}

\begin{equation}
\label{lq2_eq_phi2_update_discrete}
\begin{aligned}
\phi_{2, n+1}& \leftarrow \Pi_{K_{2}}\biggl(  \phi_{2,n} - a_n \sum_{k=0}^{\left\lfloor \frac{T}{\Delta t} -1 \right\rfloor} 
\biggl\{ \frac{\partial \log \pi}{\partial \phi_2^{-1}}\left(u_n(t_k) \mid t_k, x_n(t_k); \bm\phi_n\right) \\
&\biggl[ J\left(t_{k+1}, x_n(t_{k+1}); \bm\theta_n \right) 
- J\left(t_k, x_n(t_k); \bm\theta_n \right) - \frac{1}{2}Qx_n(t_k)^2 \Delta t \\
&+ \gamma_n {p}\left(t_k, \bm\phi_n\right) \Delta t \biggl] + \gamma_n \frac{\partial {p}}{\partial \phi_2^{-1}}\left(t_k, \bm\phi_n \right) \Delta t \biggl\} \biggr).
\end{aligned}
\end{equation}

\subsection{Pseudocode}
\label{sec:pseudocode}

Based on the analytical development presented, we outline the RL algorithm for the ALM problem as follows:

\begin{algorithm}[htb]
\caption{ALM-RL Algorithm}\label{algo_alm_rl}
\begin{algorithmic}
\For{$n = 1$ to $N$}
    \State Set $k = 0$, $t = t_k = 0$, $x_n(t_k) = x_0$
    \While {$t < T$}
        \State Sample action $u_n(t_k)$ following stochastic policy \eqref{policy_parametrization}
        \State Update next surplus deviation $x_n(t_{k+1})$ using \eqref{eq_classical_dynamics}
        \State Increment time: $t_{k+1} = t_k + \Delta t$
    \EndWhile
    \State Collect trajectory $\{(t_k, x_n(t_k), u_n(t_k))\}_{k \geq 0}$
    \State Update $\bm\theta$ and $\phi_1$ via \eqref{lq2_eq_theta_update_discrete} and \eqref{lq2_eq_phi1_update_discrete}
    \State Perform adaptive actor exploration using \eqref{lq2_eq_phi2_update_discrete}
    \State Apply scheduled critic exploration via \eqref{lq2_eq_gamma_definition}
\EndFor
\end{algorithmic}
\end{algorithm}


\section{Convergence Results}
\label{sec:theorem}
In this section, we present the convergence analysis for Algorithm~\ref{algo_alm_rl}. Throughout, we use \(c\), and its variants, to denote generic positive constants that may vary from line to line. These constants depend only on the model parameters \(A\), \(B\), \(C\), \(D\), \(Q\), \(H\), the initial condition \(x_0\), time horizon \(T\), and the predefined algorithmic hyperparameters \(c_{\gamma}\), \(U_{\bm\theta}\), \(U_1\), \(U_2\), and \(\epsilon\).


\begin{theorem}
\label{thm_convergence}
    Suppose the learning rate sequence \(\{a_n\}\) satisfies the standard conditions:
    \begin{equation}
    \label{eq_lr_assumptions}
    \sum a_n = \infty, \quad \sum a_n^2 < \infty.
    \end{equation}
    Then, the following almost surely convergence holds:
    \[
    \phi_{1,n} \xrightarrow{\text{a.s.}} \phi_1^* = -\frac{B + CD}{D^2},
    \]
    and
    \[
    \phi_{2,n} \xrightarrow{\text{a.s.}} \epsilon.
    \]
\end{theorem}

\begin{remark}
    \(\phi_1^*\) represents the \textit{oracle} value, corresponding to the explicit solution under the assumption of complete market knowledge; see Equations \eqref{eq_classical_value_function} and \eqref{eq_rl_value_function} for the optimal value \(\phi_1^* = -\frac{B + CD}{D^2}\).
\end{remark}

\begin{proof}
The proof strategy builds on the framework established in \cite[Theorem 4.1]{huang2024sublinear} and \cite[Theorem 5.1]{huang2025data}, with necessary adaptations to the current Algorithm \ref{algo_alm_rl} under ALM setting. It also leverages classical results from stochastic approximation theory \cite{robbins1951stochastic,andradottir1995stochastic,robbins1971convergence}.

Firstly, we denote the mean part \(h_1(\phi_{1,n}, \phi_{2,n}; \bm{\theta}_n) = \E[Z_{1,n}(T) \mid \bm{\theta}_n, \bm{\phi}_n]\) and noise part \(\xi_{1,n} = {Z}_{1,n}(T) - h_1(\phi_{1,n}, \phi_{2,n}; \bm{\theta}_n)\), so that the updating rule for \(\phi_1\) is
\begin{equation}
\label{eq_phi1_update_app}
\phi_{1, n+1} = \Pi_{K_{1, n+1}} (\phi_{1,n} + a_{n}[h_1(\phi_{1,n}, \phi_{2,n}; \bm{\theta}_n) + \xi_{1, n}]).
\end{equation}
Applying Ito's lemma to the process \(J\left(t, x_n(t); \bm{\theta}_n \right)\) then to \(Z_{1,n}\), we have

\begin{equation}
\label{eq_dz1_app}
\begin{aligned}
    \dd& Z_{1, n}(t) = (u_n(t)-\phi_{1,n}x_n(t))x_n(t) \biggl\{ \biggl [ -\frac{1}{2}k_1^\prime(t; \bm\theta_n)x_n(t)^2   \\
    &+ k_3^\prime(t; \bm\theta_n)- (Ax_n(t) + B u_n(t))k_1(t; \bm\theta_n)x_n(t) - \frac{1}{2}Qx_n(t)^2\\
    &- \frac{(Cx_n(t)+D u_n(t))^2}{2} k_1(t; \bm\theta_n) + \frac{\gamma}{2} \log( 2 \pi e \phi_{2,n})  \biggl] \dd t\\
    &- \biggl( (Cx_n(t)+D u_n(t)) k_1(t; \bm\theta_n) x_n(t) \biggl) \dd W_n(t)\biggr\}.\\
\end{aligned}
\end{equation}

Then by \cite[Lemma B.1]{huang2024sublinear}, we can get the noise bound
\begin{equation}
\label{eq:noise_upper1}
\begin{aligned}
&\operatorname{Var}\left( \xi_{1,n} \Big| \bm\theta_n, \phi_{1,n}, \phi_{2,n} \right) \\ \leq& c' \left( 1 +  |\phi_{1,n}|^{8} + (\log \phi_{2,n})^8\right) \exp{\{c'|\phi_{1,n}|^6\}}\\
\leq& c' \left( 1 +  U_1^{8} + (\log U_2)^8 + (\log \epsilon)^8\right) \exp{\{c'U_1^6\}} \leq c,
\end{aligned}
\end{equation}
and the mean part
\begin{equation}
\label{eq_h1_app}
\begin{aligned}
    h_1(\phi_{1,n}, \phi_{2,n}; \bm\theta_n) =-l(\phi_{1,n}, \phi_{2,n}; \bm\theta_n)(\phi_{1,n} - \phi_1^*),
\end{aligned}
\end{equation}
where
\begin{equation}
\label{eq_l_definition}
l(\phi_{1,n}, \phi_{2,n}; \bm\theta_n) = D^2 \phi_{2,n} \int_0^T k_1(t; \bm\theta_n)\E [x_n(t)^2] \dd t.
\end{equation}
Moreover, we can further derive that \(l(\phi_{1,n}, \phi_{2,n}; \bm\theta_n) \geq \Bar{c} > 0\) and \(|h_1(\phi_{1,n}, \phi_{2,n}; \bm\theta_n)| \leq c'U_2(1+U_1)e^{c'U_1^2} \leq c\).

Next, we let $\{\g_n\}$ be the filtration generated by $\{\bm\theta_m, \phi_{1,m}, \\ \phi_{2,m}, m=0,1,2,...,n\}$ and denote \(U_{1,n} = \phi_{1,n} - \phi_1^*\). Then we have 
\[\begin{aligned}
& \E\left[|U_{1,n+1}|^2 \Big| \g_n \right] \\
\leq & \E\left[| U_{1,n} +  a_n[ h_1(\phi_{1,n}, \phi_{2,n}; \bm\theta_n) + \xi_{1,n}] |^2 \Big| \g_n \right] \\
\leq & |U_{1,n}|^2 + 2a_n  U_{1,n}  h_1(\phi_{1,n}, \phi_{2,n}; \bm\theta_n) + \\
&+ 3a_n^2 \left( | h_1(\phi_{1,n}, \phi_{2,n}; \bm\theta_n)|^2 +  \E\left[ \left|\xi_{1,n} \right|^2 \Big| \g_n\right] \right) \\
\leq & |U_{1,n}|^2 + 2a_n U_{1,n}  h_1(\phi_{1,n}, \phi_{2,n}; \bm\theta_n)+ca_n^2.
\end{aligned}\]

Following from \cite[Theorem 1]{robbins1971convergence}, we know that $\left|U_{1,n}\right|^2$ converges to a finite limit and $\sum -a_n U_{1,n}  h_1(\phi_{1,n}, \phi_{2,n}; \bm\theta_n)<\infty$ almost surely.
Then by \eqref{eq_h1_app}, \eqref{eq_l_definition} and lower bound of \(l(\phi_{1,n}, \phi_{2,n}; \bm\theta_n)\),
\[
\begin{aligned}
-a_n U_{1,n}  h_1(\phi_{1,n}, \phi_{2,n}; \bm\theta_n)=2a_nl(\phi_{1,n}, \phi_{2,n}; \bm\theta_n) U_{1,n}^2\geq 2\Bar{c}a_nU_{1,n}^2.
\end{aligned}
\]
To prove $U_{1,n}^2 \rightarrow 0$, we suppose $U_{1,n}^2 \rightarrow r$ almost surely, where \(0<c<\infty\) is a constant. Then there exists an \(n_0\) and \(0<\delta<r\) such that \(U_{1,n}^2 \geq r-\delta>0\) for \(n>n_0\). Thus, by the assumption of this theorem, we have
\[
\sum -a_n U_{1,n}  h_1(\phi_{1,n}, \phi_{2,n}; \bm\theta_n) \geq \sum 2\Bar{c}a_nU_{1,n}^2 \geq \sum 2\Bar{c}a_n(r-\delta)=\infty,
\]
which contradicts with $\sum -a_n U_{1,n}  h_1(\phi_{1,n}, \phi_{2,n}; \bm\theta_n)<\infty$. Therefore, \(\phi_{1,n}\) converges to \(\phi_1^*\) almost surely. The almost sure convergence of \(\phi_{2,n}\) to \(\epsilon\) follows from a similar argument as \(\phi_{1,n}\). 
\end{proof}

This proof is included for completeness and builds on \cite{huang2024sublinear,huang2025data}, with key differences in the use of uniform bounds and a minimum exploration level specific to ALM.

\section{Numerical Experiments}
\label{sec:numerical_experiments}
This section details simulation experiments that compare our ALM-RL algorithm against six alternative strategies. The comparisons include two enhanced traditional financial methods, one model-based continuous-time RL strategy, and three established RL algorithms, each described in the subsequent subsection.

\subsection{Comparative ALM Strategies}
\label{sec:comparative_strategies}

\subsubsection{Dynamic CPPI Strategy}
\label{sec:dynamic_cppi}
To ensure comparability with our ALM-RL algorithm and other RL methods, the Dynamic Constant Proportion Portfolio Insurance (DCPPI) strategy incorporates an adaptive multiplier, traditionally constant in CPPI \citep{hakanoglu1989constant,black1992theory}. This adjustment enhances the strategy's ability to learn and adapt, overcoming the traditional CPPI's limitation where the performance is highly dependent on the initially chosen multiplier \(m\).

DCPPI seeks to maintain zero deviation between the current surplus and the target surplus by dynamically adjusting to changing market conditions. The policy \(u\) is defined as:
\begin{equation}
\label{eq_u_zero}
u^{DCPPI}(t) = -m \cdot x(t),
\end{equation}
where \(m\) is adaptively updated using a data-driven approach. Starting from an initial value \(m_0\), we simulate a trajectory of surplus deviation \(x_0, x_1, ..., x_l\), and adjust \(m\) based on the directionality of changes between consecutive surplus deviations:
\begin{equation}
\label{eq_m_update}
m_{n+1} = m_n + a_n \cdot \text{sgn}\left(\sum_{i=0}^{l-1} \text{sgn}(x_i \cdot x_{i+1})\right),
\end{equation}
where \(a_n\) is the learning rate and \(\text{sgn}(\cdot)\) is the sign function. This updating rule ensures that \(m\) is modified to correct the previous trajectory’s trend by considering the sign consistency between consecutive surplus deviations, enhancing the responsiveness and accuracy of the strategy in aligning with market dynamics.

\subsubsection{Adaptive Contingent Strategy}
\label{sec:adaptive_contingent}
Drawing on the principles of contingent immunization \citep{leibowitz1982contingent,leibowitz1983contingent}, the Adaptive Contingent Strategy (ACS) aims to maintain the surplus deviation within predefined tolerance levels (\(\delta\)), contrasting with the DCPPI's zero-deviation from the target. This conservative approach allows for minor fluctuations within a safe boundary and opts for inaction when the surplus deviation is adequately balanced, thereby avoiding unnecessary market exposure and reducing noise amplification from market volatility. This is particularly useful given the stochastic nature of financial movements, specifically the volatility component \(D u(t) \mathrm{d}W(t)\) in the ALM dynamics \eqref{eq_classical_dynamics}.

The policy \(u\) is formulated to be minimally interfered:
\begin{equation}
\label{eq_u_bound}
u^{ACS}(t) = -m \cdot \text{sgn}(x(t)) \cdot \max(|x(t)| - \delta, 0),
\end{equation}
where the multiplier \(m\) dynamically updates similarly to \eqref{eq_m_update}.

\subsubsection{Model-Based Plugin Strategy}
\label{sec:model_based_plugin}
The Model-Based Plugin Strategy (MBP), derived from the continuous-time RL algorithm by \cite{basei2022logarithmic,szpruch2024optimal}, primarily estimates parameters \(A\) and \(B\) under assumptions of constant volatility, and then plugs these estimates into analytical solutions. This algorithm has been mathematically proven to offer fast convergence. To align with the dynamic complexities of the ALM problem, which involves state- and control-dependent volatilities, this approach has been extended to also estimate parameters \(C\) and \(D\) using least squares regression, as detailed in \cite{huang2024sublinear}. This extension provides a clear contrast to our model-free, continuous-time RL approach.

\subsubsection{Advanced RL Strategies}
\label{sec:advanced_rl_strategies}
In our comparative analysis, we include three prominent RL algorithms—Soft Actor-Critic (SAC) \citep{haarnoja2018soft}, Proximal Policy Optimization (PPO)\citep{schulman2017proximal}, and Deep Deterministic Policy Gradient (DDPG) \citep{lillicrap2015continuous}—due to their distinct characteristics and relevance in advancing RL applications. SAC is selected for its entropy-enhanced exploration technique which aligns with our model's entropy-based approach, emphasizing efficient exploration in continuous action spaces. PPO is included as a state-of-the-art representative for its ability to ensure stable and reliable policy updates, which is crucial for consistent performance across diverse market conditions. DDPG is chosen as a commonly referenced benchmark in RL studies, known for its foundational role in integrating deterministic policy gradient concepts with deep learning frameworks.

\subsection{Experiment Setup}  
\label{sec:setup}  
To evaluate the performance of our method, we conduct simulations designed to reflect realistic and uncertain ALM scenarios. While most existing studies in the ALM literature \citep{li2013time,chiu2012mean,yao2013continuous,zhang2016mean} evaluate algorithms under fixed model parameters, we adopt a randomized setup to better reflect the lack of prior knowledge in real-world financial markets and to assess the algorithm's robustness across diverse environments. The parameter ranges are chosen based on typical values used in these studies and general financial intuition: \(A \sim \mathcal{U}(-0.05, 0.05)\), \(B \sim \mathcal{U}(0.05, 0.15)\), and \(C, D \sim \mathcal{U}(0.1, 0.2)\). Each simulation runs for 20000 episodes with a discretization step size of \(\Delta t = 0.01\), and is repeated independently 200 times to ensure statistically reliable results.

Furthermore, the learning rate \(a_n = (n+1)^{-3/4}\), shown to be effective in \cite{huang2024sublinear}, is used for ALM-RL as well as for the two enhanced traditional ALM methods, DCPPI and ACS. This choice satisfies the standard assumption in~\eqref{eq_lr_assumptions}, which is also the only assumption required in Theorem~\ref{thm_convergence}. For ALM-RL, the exploration scheduling sequence \(b_n = (n+1)^{1/4}\) is additionally employed to accelerate convergence, as demonstrated in~\cite{huang2024sublinear}, from which both \(a_n\) and \(b_n\) are adopted. The initial actor exploration level is set to \(\phi_{2,0} = 1\), and the constant \(c_{\gamma} = 1\). The projection bounds are set to \(U_{\bm\theta} = U_1 = U_2 = 100\), and the minimum exploration level is \(\epsilon = 0.01\). The tolerance level for ACS is set as \(\delta=0.1\). Finally, all other learnable parameters across all ALM strategies are initialized using standard normal distributions.

The settings for MBP follow those in \cite{basei2022logarithmic,szpruch2024optimal,huang2024sublinear}. For SAC, PPO, and DDPG, the neural networks (NNs) used for both the actor and critic are feedforward architectures with two hidden layers, each containing 32 neurons and ReLU activation. Other hyperparameters are mostly adopted from \cite{haarnoja2018soft,schulman2017proximal,lillicrap2015continuous}. Lastly, to ensure reproducibility and fair comparisons under the randomized environment, we use 200 different random seeds to represent 200 independent market scenarios. Each method is evaluated under the same set of seeds so that all strategies face identical market conditions in each run.

\subsection{Evaluation Metric}

The average reward, a commonly used metric in RL to assess performance, is employed to compare the effectiveness of our ALM-RL algorithm with six alternative strategies. Following the value function \eqref{eq_classical_lq}, for each independent experiment, the reward is computed as:

\begin{equation}
\text{Reward} = \sum_{k=0}^{\left\lfloor \frac{T}{\Delta t} -1 \right\rfloor} -\frac{1}{2} Q \left(x_n(t_k)\right)^2 \Delta t - \frac{1}{2} H \left(x_n(T)\right)^2,
\end{equation}
where \(x_n(t_k)\) represents the surplus deviation at time step \(t_k\).

For each method, the average reward per episode is computed as the mean of rewards from 200 independent experiments, resulting in an average reward curve over 20,000 episodes. This curve provides a reliable measure for comparing the methods' performance, learning dynamics, and overall effectiveness.

\subsection{Performance Evaluation of ALM Strategies}
\label{sec:performance_analysis}

\begin{figure}[h]
  \centering
  \includegraphics[width=0.98\linewidth]{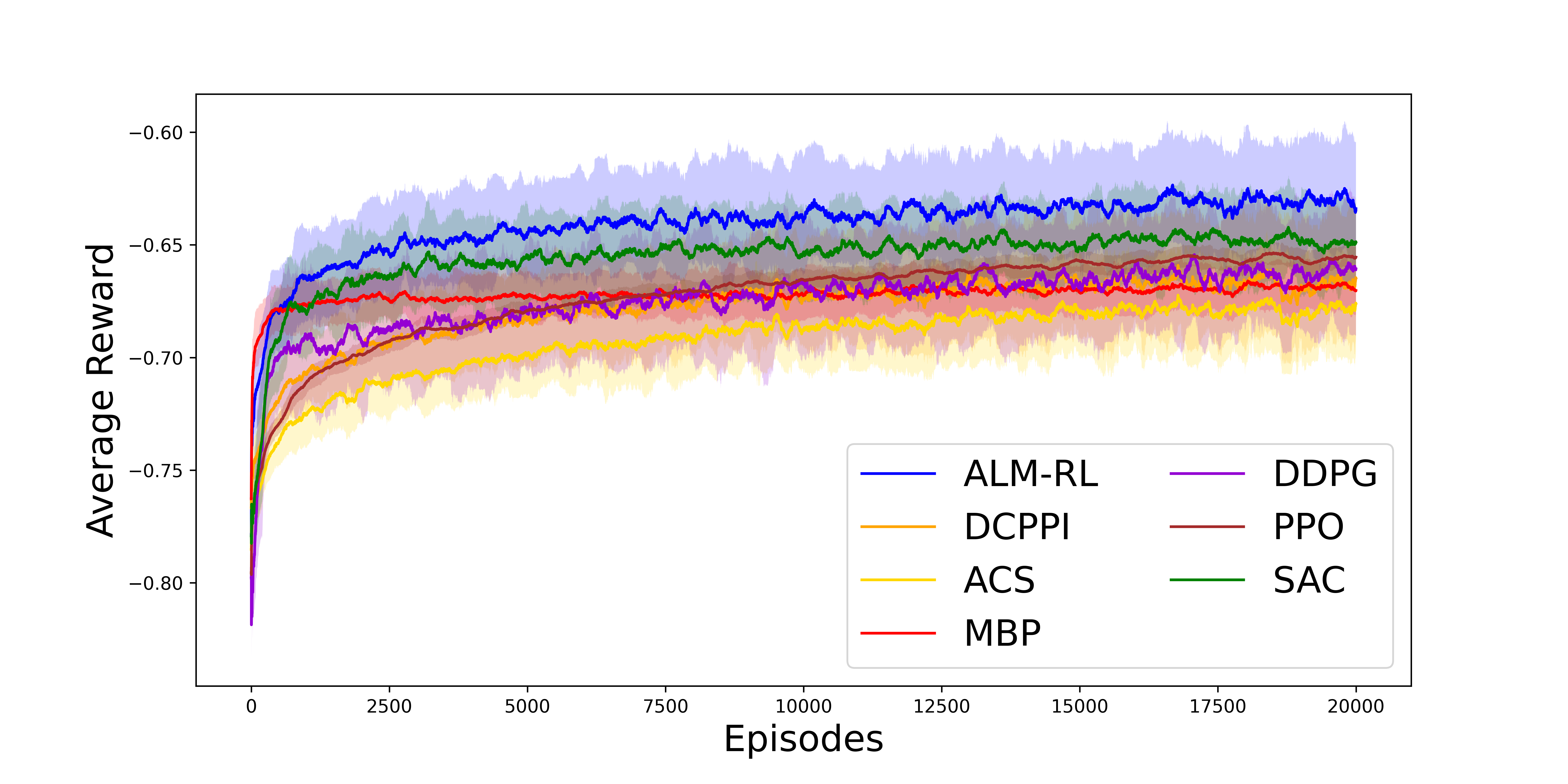}
  \caption{Average reward under randomized market parameters, smoothed with a 200-point moving average over 20000 episodes. The shaded area indicates the interquartile range, based on 200 independent simulations.}
  \label{fig:vf_comparison}
\end{figure}

Now we analyze the performance of ALM strategies in randomized market conditions across 200 independent runs, as illustrated in Figure \ref{fig:vf_comparison}.

From Figure~\ref{fig:vf_comparison}, we see that our proposed ALM-RL algorithm consistently outperforms all other strategies across almost all episodes. It exhibits rapid initial gains and sustained superiority throughout the learning horizon, demonstrating notable resilience and adaptability in refining its strategy more effectively than competing methods. This strong performance is likely due to its use of entropy-regularization exploration techniques and non-degenerate stochastic policies, which enhance adaptability and decision-making under uncertainty. Similarly, SAC, which also leverages entropy techniques and stochastic policies, achieves rapid initial gains and maintains the second-best performance after 2500 episodes. 

PPO, due to its conservative clipped surrogate objective, starts off slower but gradually approaches SAC’s performance, albeit remaining slightly lower. The clipping mechanism yields the smoothest learning curve and the narrowest interquartile range (IQR) among all strategies, reflecting high stability across runs. Compared with SAC and PPO, DDPG exhibits moderate initial growth but ultimately settles at a lower performance level. It also exhibits the highest volatility and widest IQR, likely due to the sensitivity of its deterministic policy gradient to noise and outliers. The MBP strategy demonstrates rapid early growth and attains a relatively high average reward in the initial stage. However, its performance soon stagnates, converging to suboptimal solutions. This lack of continued improvement is reflected in the flat reward curve and is likely attributable to parameter estimation errors inherent in financial markets~\cite{luenberger1998investment}.
Finally, DCPPI and ACS achieve one of the lowest terminal rewards but maintain reliable, smooth performance throughout. Moreover, DCPPI’s proactive control yields better outcomes than the more conservative ACS, reinforcing the importance of active management.

\begin{figure}[h]
  \centering
  \includegraphics[width=0.98\linewidth]{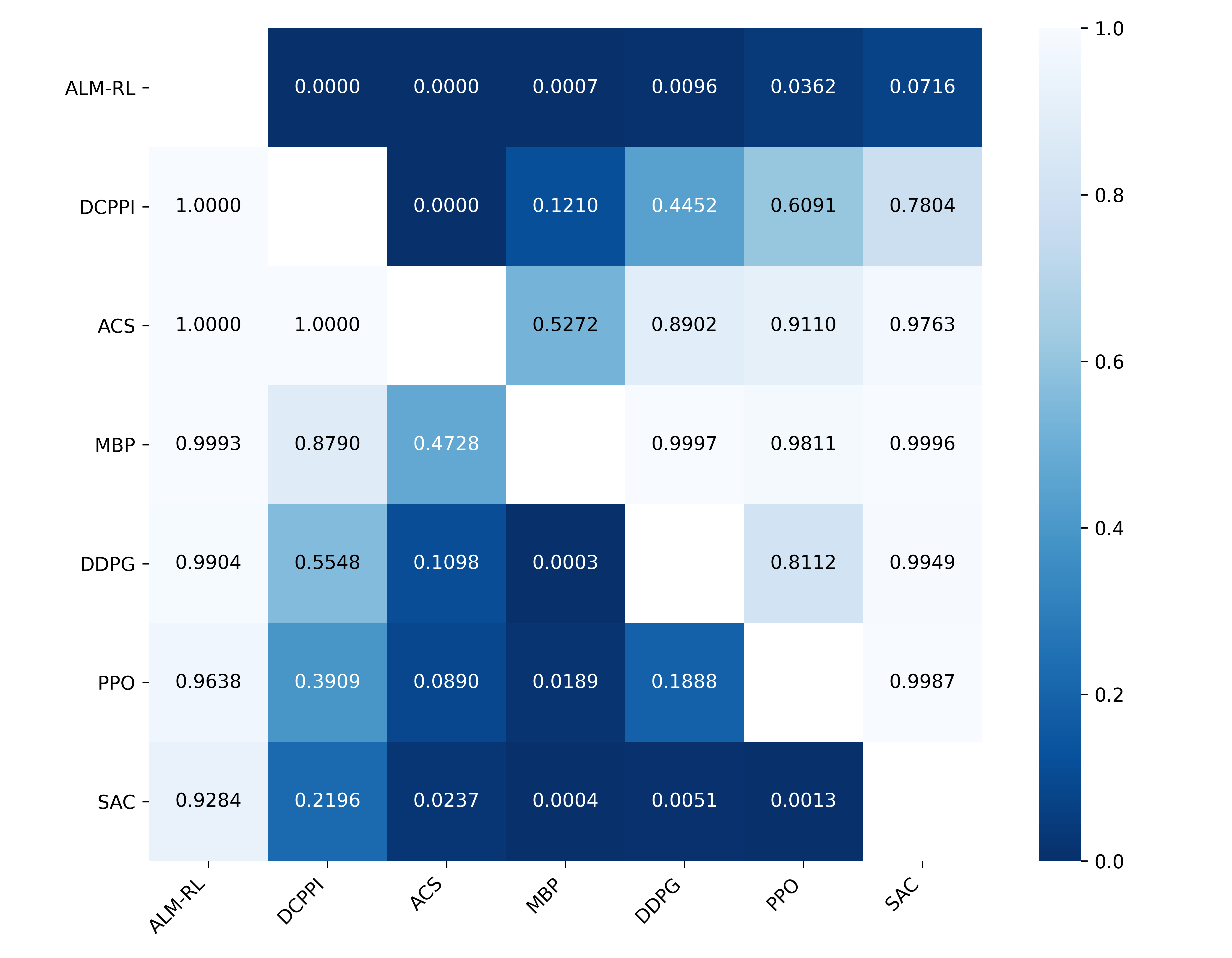}
  \caption{Heatmap of $p$-values from one-sided Wilcoxon paired tests comparing the terminal reward of each ALM strategy. The terminal reward is calculated as the average reward over the last 500 episodes to reduce the noise that may result from using a single final episode.}
  \label{fig:pvalues}
\end{figure}

Moreover, in order to show statistical significance, we conduct one-sided Wilcoxon paired tests between each pair of ALM strategies, and the resulting matrix of $p$-values is presented in Figure~\ref{fig:pvalues}. Each cell displays the $p$-value for the null hypothesis that the row method does not outperform the column method. Darker shades indicate stronger statistical evidence against the null. Notably, ALM-RL demonstrates statistically significant improvements over all strategies except SAC at the 95\% confidence level, and over SAC at the 90\% level, with corresponding $p$-values below 0.05 and 0.10, respectively. This supports the robustness and consistent superiority of ALM-RL across randomized environments.

\section{Conclusions}
\label{sec:conclusions}

This paper introduced a novel approach for ALM by formulating the problem as a stochastic LQ control and solving it within a model-free, continuous-time RL framework. In addition to the new formulation, we integrated adaptive exploration for the actor and scheduled exploration for the critic, ensuring an effective exploration and exploitation trade-off. Notably, despite the use of adaptive and scheduled exploration techniques, we are able to prove almost sure convergence of all policy parameters. Our policy gradient-based soft actor-critic method was evaluated against two enhanced traditional financial strategies, a model-based continuous-time RL approach, and state-of-the-art RL algorithms, including SAC, PPO, and DDPG. The results consistently demonstrate that our method outperforms these alternatives across diverse market conditions.

The superior performance results from directly learning optimal ALM strategies without assuming any knowledge of the financial environment or estimating market parameters, highlighting a fundamental advantage of our approach. Future research will focus on extending this framework to broader financial domains and evaluating its performance in more complex and dynamic market environments.

\bibliographystyle{ACM-Reference-Format}
\bibliography{icaif25}






\end{document}